\documentclass[11pt,a4paper]{article}
\usepackage{arabtex}
\usepackage{utf8}
\usepackage[hyperref]{acl2017}

\usepackage{graphicx}
\usepackage{enumerate}

\usepackage{times}
\usepackage{latexsym}
\usepackage{url}
\usepackage{verbatim}
\usepackage{booktabs}
\setcode{utf8}

\aclfinalcopy

\title{Robust Tuning Datasets for Statistical Machine Translation}

\author{Preslav Nakov \and Stephan Vogel\\
ALT Research Group\\
Qatar Computing Research Institute, HBKU \\
  {\tt \{pnakov, svogel\}@hbku.edu.qa}}

\date{}

\begin{document}
\maketitle
\begin{abstract}
We explore the idea of automatically crafting a tuning dataset for Statistical Machine Translation (SMT) that makes the hyper-parameters of the SMT system more robust with respect to some specific deficiencies of the parameter tuning algorithms.
This is an under-explored research direction, which can allow better parameter tuning. 
In this paper, we achieve this goal by selecting a subset of the available sentence pairs, which are more suitable for specific combinations of optimizers, objective functions, and evaluation measures.
We demonstrate the potential of the idea with the pairwise ranking optimization (PRO) optimizer, which is known to yield too short translations.
We show that the learning problem can be alleviated by tuning on a subset of the development set,
selected based on sentence length.
In particular, using the longest 50\% of the tuning sentences,
we achieve two-fold tuning speedup,
and improvements in BLEU score that rival those of alternatives,
which fix BLEU+1's smoothing instead.
\end{abstract}

\section{Introduction}
\label{sec:intro}
Modern Statistical Machine Translation (SMT) systems have several, somewhat independent, 
components that work together to generate a good translation,
and are typically combined in a log-linear framework,
where the language model, the translation model, the reordering model, etc.,
contribute to the hypothesis score with different weights.
It is now standard to learn these weights discriminatively, from a development dataset,
e.g.,~by optimizing BLEU \cite{Papineni:Roukos:Ward:Zhu:2002} or some other measure directly.

\noindent A tuned system can often yield very significant improvements in terms of translation quality
compared to a system that uses standard, untuned default parameters.
Thus, a lot of research attention in SMT has been devoted
to designing different algorithms for parameter optimization.
For years, it was typical to use minimum error rate training, or MERT \cite{och03minimum},
which works quite well when the number of parameters is small.
As the number of parameters has grown,
rivaling optimizers such as MIRA \cite{watanabe-EtAl:2007,Chiang:2008} and PRO \cite{Hopkins:2011}
have been developed, as well as various variations thereof \cite{bazrafshan-chung-gildea:2012:NAACL-HLT,cherry-foster:2012:NAACL-HLT,gimpel-smith:2012:NAACL-HLT1}.

These optimization algorithms have focused on learning from a given fixed dataset,
relying on the standard machine learning 
assumption that the training and the development data come from the same distribution 
as the test data, e.g., in terms of domain, coverage, genre, length, etc.
In practical terms, 
this is especially important for the development/tuning data, 
but with standard datasets, there is often no way to guarantee this,
and many researchers have determined empirically the most suitable tuning dataset
by observing the translation score on the test dataset.
Yet, the choice of tuning dataset can considerably affect the results;
for example, \newcite{Zheng2010} report variation across different standard NIST tuning MTxx datasets of over six BLEU points
for Chinese-English SMT
when testing on the NIST MT08 test dataset. 

Given a \emph{reasonable} tuning set, i.e., one 
that is really 
coming from the same distribution as the test dataset, a \emph{good} optimization algorithm should be able to learn to produce optimal weights.
Yet, the choice of optimization objective can yield dramatically different translations
since different algorithms might need to stress some aspects of the tuning dataset and downplay others.

\noindent One reason for this is that different optimizers interact differently with different objectives.
For example, we have previously shown that sentence-level optimizers yield too short translations when optimizing BLEU+1 \cite{nakov-guzman-vogel:2012:PAPERS}.

In this paper, we advocate 
the idea of 
automatically crafting a tuning dataset that makes tuning parameters \emph{less} susceptible to the deficiencies of the learning algorithms.
More specifically,
in order to bridge this gap,
we propose to customize the tuning dataset by selecting a subset of the available sentence pairs,
taking into account the target domain and the peculiarities of the optimization algorithm,
objective function, and evaluation measure used.
 This is important because it brings us a step closer to robust learning, instead of simply overfitting the tuning dataset. 

Below, we focus specifically on sentence length as a selection criteria. We chose length because it can have consequences on how translations are scored w.r.t. to metrics like BLEU, and besides, it is a known issue for PRO.  
Still, to the best of our knowledge,
the interaction between the optimizer,
the optimization objective, the evaluation measure,
and the development dataset has been largely neglected so far.
For instance, we show that the problem of short translations when tuning with PRO \cite{Hopkins:2011}
is worsened when tuning on short sentence pairs, and alleviated when
emphasizing the longer tuning sentences.

Tuning set crafting can be done in various ways, 
e.g.,
by removing examples with sub-optimal characteristics (e.g., short sentences)
or by oversampling the ones with desired characteristics (e.g., longer sentences).
Here we focus on selection from a single tuning set because it is applicable to different datasets.
We show that significant performance gains are possible with PRO
when selecting just a subset of the tuning dataset based on length.
Our objective here is to draw the attention of the research community to the possibilities that dataset customization through subset selection can offer
for different experimental conditions.
We believe that this is a very promising, yet largely underexplored 
research direction.

Naturally, one could also try to select data from elsewhere and build a completely custom dataset, e.g., by selecting sentences from the training dataset. However, this is not possible in case of multiple references (the training bi-text has only one reference).

\noindent One could also try to select/fuse from different available tuning datasets,
but it is rare to have multiple tuning datasets.

It has also been observed that having multiple references in the tuning dataset
can yield more accurate parameter estimates and thus better test translation scores \cite{madnaniamta08}.
Thus, adding a few human translations, could significantly boost translation quality.
Since this is costly, some researchers have resorted to using automatically generated references,
with modest performance gains.

The remainder of the paper is organized as follows:
Section~\ref{sec:related} introduces related work,
Section~\ref{sec:method} describes the method,
Section~\ref{expsetup} presents the experimental setup,
Section~\ref{sec:eval} discusses the evaluation results,
and
Section~\ref{sec:discuss} provides deeper analysis and
further discussion.
Finally, Section~\ref{sec:conclusion}
concludes with possible
directions for future work.

\section{Related Work}
\label{sec:related}

Tuning the parameters of a log-linear model for SMT is an active area of research.
The typical way to do this is to use minimum error rate training, or MERT, \cite{och03minimum},
which optimizes the standard dataset-level BLEU directly.

Recently, there has been a surge in new optimization techniques for SMT.
Most notably, this includes
the margin-infused relaxed algorithm or MIRA \cite{watanabe-EtAl:2007,Chiang:2008,Chiang:2009},
which is an on-line sentence-level perceptron-like passive-aggressive optimizer,
and pairwise ranking optimization or PRO \cite{Hopkins:2011},
which operates in batch mode and sees tuning as ranking.

A number of improved versions thereof have been proposed
including a batch version of MIRA \cite{cherry-foster:2012:NAACL-HLT}
and a linear regression version of PRO \cite{bazrafshan-chung-gildea:2012:NAACL-HLT}.
Another recent optimizer is Rampeon \cite{gimpel-smith:2012:NAACL-HLT1}.
We refer the interested reader to three recent overviews on parameter optimization for SMT:
\cite{McAllester:Keshet:2011,cherry-foster:2012:NAACL-HLT,gimpel-smith:2012:NAACL-HLT1}.

With the emergence of new optimization techniques, there have been also studies that compare stability
between MIRA--MERT \cite{Chiang:2008,Chiang:2009,cherry-foster:2012:NAACL-HLT},
PRO--MERT \cite{Hopkins:2011},
MIRA--PRO--MERT \cite{cherry-foster:2012:NAACL-HLT,gimpel-smith:2012:NAACL-HLT1,nakov-guzman-vogel:2012:PAPERS}.

\noindent More relevant to the present work,
there has been some interest in analyzing how different optimizers interact with specific metrics.
For example, pathological verbosity was reported when tuning MERT on recall-oriented metrics such as METEOR \cite{Lavie:2009:MMA,denkowski2011meteor},
large variance was observed with MIRA \cite{simianer2012joint},
and \emph{monsters} were found when using PRO with too long tuning sentences \cite{nakov-guzman-vogel:2013:Short}.
In previous work, we also found that MERT learns verbosity, while PRO learns length \cite{guzman-nakov-vogel:2015:CoNLL}.

It has been also observed that having multiple references for the tuning dataset
can yield 
better test-time translation performance \cite{madnani-EtAl:2007:WMT}.
Thus, adding a few extra human reference translations could significantly boost translation quality.
Since this is costly, some researchers have resorted to using automatically generated references,\footnote{Paraphrasing was also applied to the training bi-text \cite{Nakov:2008:ISM,nakov-ng:2009:EMNLP} and to the phrase table \cite{Callison-Burch:al:2006:mt}.}
via paraphrasing \cite{madnaniamta08} and back-translation \cite{dyer-EtAl:2011:WMT},
with modest performance gains.

There has been also work on tuning data selection and/or fusion
in the special case when multiple versions of the source sentence are available \cite{nakov-EtAl:2013:RANLP-2013}.
This is a fairly rare situation for such approaches to be broadly applicable.

Most relevant to our work,
there were efforts to build tuning datasets using
information retrieval \cite{Zheng2010,tamchyna-EtAl:2012:WMT},
text clustering \cite{li-EtAl:2010:PAPERS3},
and sentence-length based features \cite{guzman-EtAl:2012:WMT}.
To avoid data sparseness, most of these approaches require a larger pool of data,
which they typically select from the training bi-text,
thus reducing the amount of data available for model training.
This makes such approaches inapplicable in multi-reference testset contexts
since the bi-text only has one translation per source sentence.
Moreover, the selection is typically done based on the actual test input,
which is not known a priori in a realistic SMT setup, e.g.,~in online translation.

In contrast, we aim to produce customized datasets 
that 
are less susceptible to that, and
are 
suitable for specific combinations of optimizers, objective functions, and evaluation measures.
This can yield better parameter estimation, while using less data
and
more efficient tuning with faster iterations and a smaller computational footprint.






\section{Method}
\label{sec:method}
Below we present one particular example of tuning dataset customization
in order to illustrate the potential of the idea.

In previous work, we have shown that the PRO optimizer yields SMT parameters
that yield test-time translations that are shorter than they should be.
We have addressed this by changing the objective function, sentence-level BLEU+1,
and we have proposed to replace it with 
one with better smoothing \cite{nakov-guzman-vogel:2012:PAPERS}.
Here we propose an alternative solution,
which customizes the tuning dataset by selecting a subset of higher average length.

Observe that,
if the reason for PRO yielding too short translations is the add-one smoothing in BLEU+1,
this should affect shorter sentences to a greater extent,
since the effect of the smoothing is bigger for them.
I.e., we should expect that, when tuning with PRO,
the translations of short sentences 
should get relatively shorter translations
than those of long sentences.
This means that we should expect to get longer translations if we tune on longer sentences,
i.e., if we customize the tuning dataset,
which can be done, e.g.,
(\emph{a})~by excluding some of the short sentences
or
(\emph{b})~by oversampling some of the long sentences.
We will explore approach (\emph{a}) below:
in particular, 
we will exclude half of the sentences, keeping the longest 50\% only.

\section{Experimental Setup}
\label{expsetup}

We experimented with Arabic-to-English SMT, training on the Arabic-English data
that was made available for the NIST 2012 OpenMT Evaluation.\footnote{\texttt{www.nist.gov/itl/iad/mig/openmt12.cfm}}
We used all training data except for the UN corpus,
we tuned on MT06 (and subsets thereof),
and we tested on MT09,
which have four English reference translations.

We trained a phrase-based SMT model \cite{Koehn:2003:SPT} as implemented in the Moses toolkit \cite{Moses:2007}.
We tokenized and truecased the English side of the training/development/testing bitexts,
and the monolingual data for language modeling using the standard tokenizer of Moses.
We segmented the words on the Arabic side of all bitexts using the MADA ATB segmentation scheme \cite{MADA}.

\begin{table}[tbh]
    \small
    \centering
    \begin{tabular}{llcc}
      \toprule
      & \bf Tuning & \bf BLEU & \bf BP \\\midrule
      1 & BP-smooth=1, grounded & 47.61 & 0.991\\
      2 & BP-smooth=1 & 47.52 & 0.984\\
        \midrule
      3 & top50 & \bf 47.47 & \bf 0.980\\
      4 & mid50 & 47.44 & 0.977\\
      5 & rand50 & 47.43 & 0.978\\
      6 & low50 & 46.38 & 0.961\\
        \midrule
      7 & full & \bf 47.18 & \bf 0.972\\
      \bottomrule
    \end{tabular}
    \caption{\label{tab:multiref}
    Multi-reference PRO experiments:
    testset BLEU and BP when tuning on different length-based subsets
    of the tuning dataset (lines 3-6).
    For comparison, we also show the results when tuning on the full tuning dataset (line 7),
    as well as what the PRO-fixes proposed in \cite{nakov-guzman-vogel:2012:PAPERS}
    would achieve when tuning on the full dataset (lines 1-2).}
  \end{table}

\noindent We then built a phrase table using the Moses pipeline with max-phrase-length 7 and Kneser-Ney smoothing,
as well as a lexicalized reordering model \cite{IWSLT:2005}: \emph{msd-bidirectional-fe}.
We used a 5-gram language model trained on GigaWord v.5 with Kneser-Ney smoothing using KenLM \cite{kenlm}.
On tuning and testing, we dropped the unknown words.

For tuning, we used PRO.
In order to avoid instabilities when tuning on long sentences,
we used a slightly modified, fixed version of PRO, as we recommended in \cite{nakov-guzman-vogel:2013:Short},
where we limited the difference between the positive and the negative example in a training sentence pair
to be no more than ten BLEU+1 points.
Moreover, in order to ensure convergence, we let PRO run for up to 25 iterations (default: 16);
we further used 1000-best lists in each iteration (default: 100).

In our experiments, we performed three reruns of parameter optimization,
and we report BLEU averaged over the three reruns, as suggested by \newcite{clark-EtAl:2011:ACL-HLT2011}
as a way to stabilize MERT.
We calculated BLEU using NIST's scoring tool v.13a, in case-sensitive mode.

\begin{table}[tbh]
    \small
    \centering
    \begin{tabular}{llcc}
      \toprule
      & \bf Tuning & \bf BLEU & \bf BP \\\midrule
      1 & BP-smooth=1, grounded & 29.68 & 0.979\\
      2 & BP-smooth=1 & 29.43 & 0.962\\
        \midrule
      3 & top50 & \bf 29.51 & \bf 0.969\\
      4 & mid50 & 29.11 & 0.950\\
      5 & rand50 & 28.96 & 0.941\\
      6 & low50 & 27.44 & 0.894\\
        \midrule
      7 & full & \bf 28.88 & \bf 0.934\\
      \bottomrule
    \end{tabular}
    \caption{\label{tab:singleref}
    Single-reference PRO experiments:
    testset BLEU and BP when tuning on different length-based subsets
    of the tuning dataset (lines 3-6).
    We also show the results when tuning on the full tuning dataset (line 7),
    as well as what the PRO-fixes proposed in \cite{nakov-guzman-vogel:2012:PAPERS}
    would achieve when tuning on the full dataset (lines 1-2).}
  \end{table}

\section{Experiments and Evaluation}
\label{sec:eval}

In this section, we verify experimentally whether tuning on short sentences can make PRO's length issue worse
and whether tuning on longer sentences could help in that respect.
We further compare the effect of tuning on long sentences
(i.e.,~of tuning dataset customization)
to using better smoothing for BLEU+1
(as we have proposed in our earlier work).

\noindent Lines 3-6 in Table~\ref{tab:multiref} show the results
when tuning on the longest (top50), middle (mid50), random (rand50) and shortest (low50)
50\% of the tuning sentences.
Comparing this to line 7 (tuning on the full MT06),
we can see that tuning on the shortest sentences
lowers the hypothesis-to-reference ratio (BP),
while tuning on top50 improves it,\footnote{The ideal target value for BP is 1.}
with the BP for mid50 and rand50 
in between.

Lines 3-6 further show that better BP 
corresponds to better BLEU.
We can also see that both BP and BLEU for 
top50
are better than those for the full MT06 tuning dataset.
Despite top50 being tuned on less data, its BP and BLEU are comparable
to those achieved by the BLEU+1 smoothing approaches shown
in lines 1-2 \cite{nakov-guzman-vogel:2012:PAPERS}, which use the full tuning dataset.

Note that when calculating BP and BLEU, for the 4-reference MT06 dataset,
we used the length of the reference 
sentence that is 
closest to the length of the hypothesis.
This is the \emph{effective reference length} from the original paper on BLEU \cite{Papineni:Roukos:Ward:Zhu:2002},
and
it is also 
the default in NIST scoring tool v13a, which we use.

Using the closest reference yields a very forgiving BP. Yet, few datasets have multiple references.
Thus, we  also experimented with a single (ref0) reference for both tuning and testing.
The results are shown in Table~\ref{tab:singleref}.
Comparing the corresponding lines of Tables \ref{tab:singleref} and \ref{tab:multiref},
we see that in this case, the length problem is more severe and affects BLEU more.
More importantly, note that the top50 customized tuning 
dataset is much more effective with a single reference translation.


\section{Discussion}\label{sec:discussion}
\label{sec:discuss}
In this section, we perform further analysis,
in order to better understand the improvements when tuning in longer sentences.
We consider three aspects:
(\emph{i})~amount of training data,
(\emph{ii})~genre overlap between tuning and test datasets,
and
(\emph{iii})~differences in the learned SMT parameters.



\subsection{Amount of Tuning Data}

\begin{figure}
  \centering
  \includegraphics[width=\linewidth]{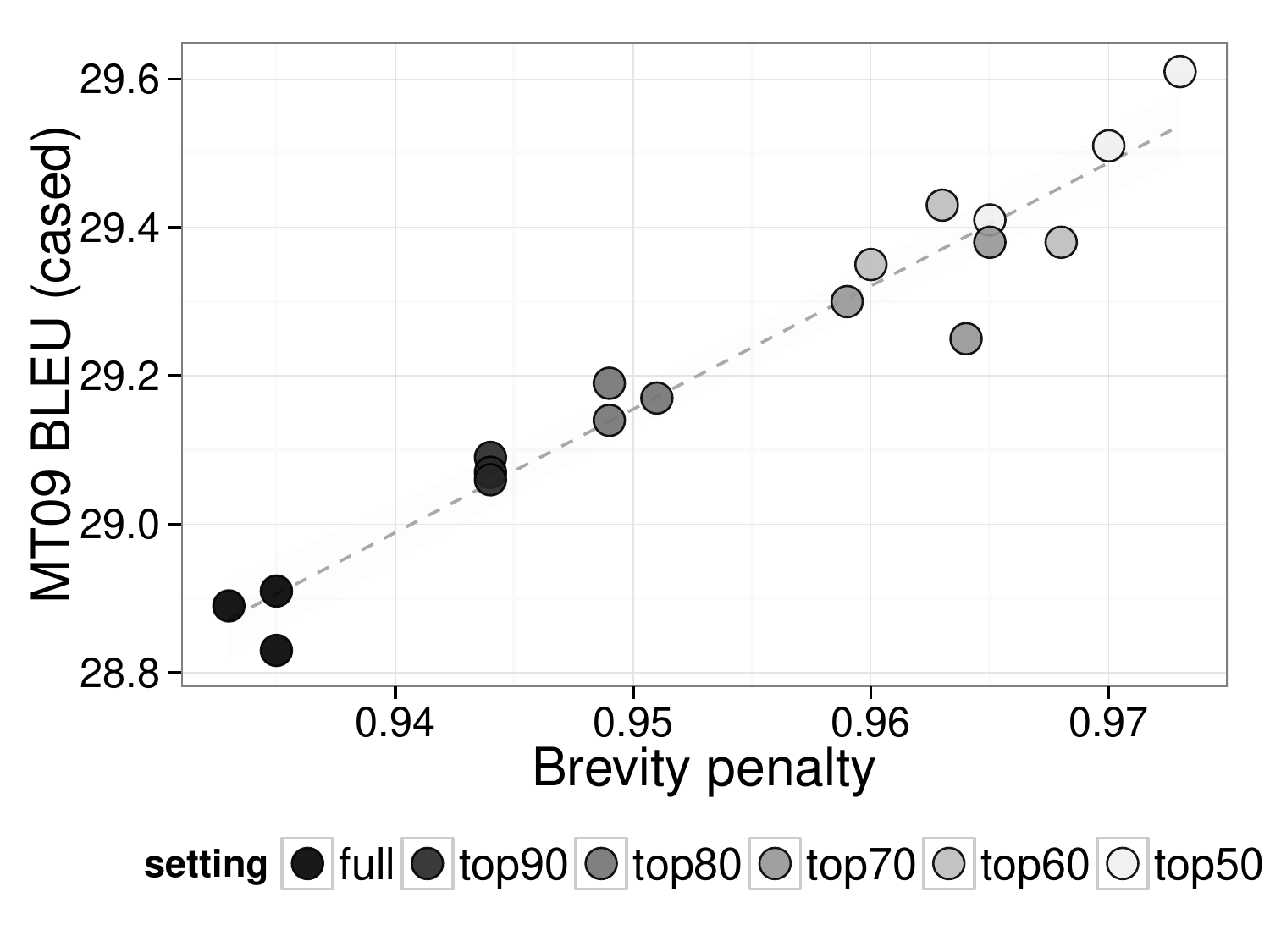}
  \caption{\label{fig:BLEU-BP}Single-reference PRO experiments.
          Correlation between cutoff, BP and BLEU score.
          The $x$-axis shows the brevity penalty (BP),
          and the $y$-axis contains the BLEU score on the testing MT09 dataset.
          Different colors show the different levels of cutoff.
          We show the results for three reruns in each setting.}
\end{figure}

In our experiments, we saw that tuning on longer sentences yields better results than tuning on shorter ones. However, one might argue that the subsets with longer sentences have access to more training data in terms of number of word tokens.
In order to shed some light on this,
we experimented with varying the percentage of longest sentences that we keep in decreasing order:
from the full dataset (100\%), we gradually removed the shortest sentences in increments of 10\%
until we ended up with just 50\% of the data.
The results are shown in Figure~\ref{fig:BLEU-BP}.
We can see that as the cutoff increases, so does BP, which in turn yields better BLEU.
This suggests that by varying the length of the tuning sentences,
we can effectively control the verbosity that PRO learns.
We can further conclude that it is not the amount of tuning data that matters but rather its characteristics.

\subsection{Genre Overlap}

MT06 is a mixture of three genres:
newswire (nw), weblogs (wb) of almost equal sizes,
and a much smaller size of broadcast news (bn).
MT09 is also a mixture, but of two genres only: it only contains newswire and weblogs.
Thus, one could ask the question of whether the observed improvements are due to better overlap
between the genres of the tuning and of the testing datasests.

\begin{table}[hbt]
\small
\centering
\begin{tabular}{ccccc}
\hline
& \bf{bn}	& {\bf nw}& {\bf wb} & $D_{KL}$ \\
\hline
{\bf MT06} \\
full	&8\%	&{\bf 46}\%	&{\bf 46}\%&3.93\\
low50	&7\%	&25\%	&64\% &7.42\\
mid50	&9\%	&{\bf 46}\%	&{\bf 41}\%&6.41 \\
top50	&8\%	&63\%	&26\% &12.09 \\
\hline		
{\bf MT09}\\		
full& &{\bf 45}\%	&{\bf 55}\%& \\
\hline
\end{tabular}
\caption{\label{tab:genres} Distribution of genres for the different partitions of the tuning data (MT06) and the test data (MT09). While MT06 has newswire (nw), and weblogs(wb) in equal amounts (with a lower proportion of broadcast news (bn)), MT09 has slightly higher proportion of weblog data than newswire. Based on Kullback--Leibler divergence ($D_{KL}$), the full partition is closest to the test data, followed by the mid50 partition.}
\end{table}

Table~\ref{tab:genres} could help answer this question;
it shows the genre distribution for the different partitions of the tuning dataset.
We can see that the distribution of genres in the full MT06 dataset is better than for top50, mid50, and low50, having the smallest Kullback--Leibler divergence with the test set;
the mid50 partition comes second.
In contrast, top50, the best-performing partition among the ones we explored,
has the most divergent genre-distribution with respect to the test dataset.
From this, we can conclude that the improvements in BLEU for top50 are definitely not due to better genre/domain overlap.


\subsection{Better Parameters}

The last question we address in this section is the following: Where exactly is the difference in translation quality coming from?
I.e., are the improvements in length only the result of decrease in the word penalty
or are there other parameters that are being affected?
In order to answer this question, we analyzed the optimized weights (averaged over three reruns) for tuning at different cutoffs, from 100\% to 50\% of the longest sentences in the MT06 development dataset.

\begin{figure}
  \centering
  \includegraphics[width=\linewidth]{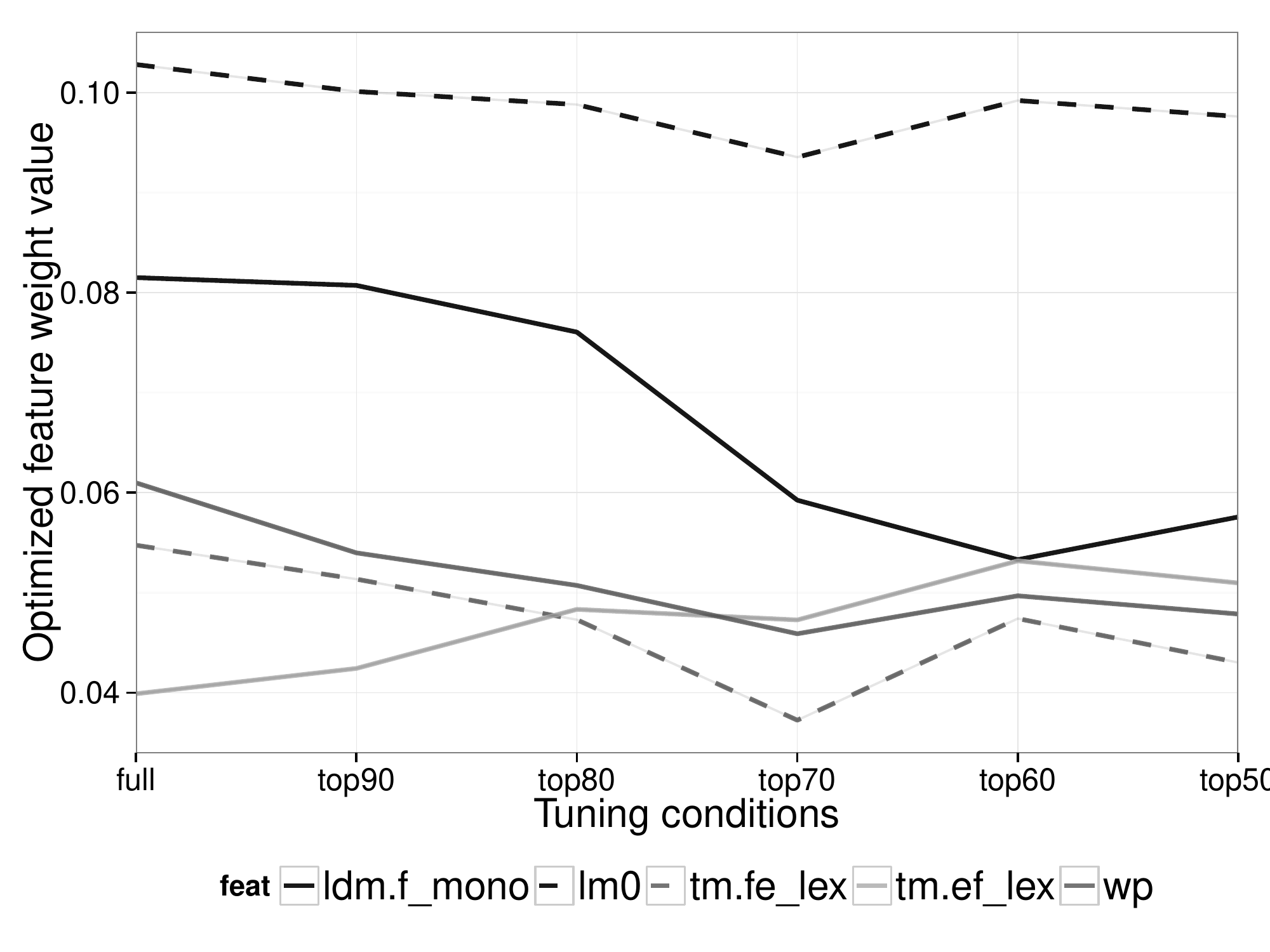}
  \caption{\label{fig:weightsel} Optimized feature weight values for each of the different tuning settings. Only a subset out of the 14 different tuning weights that are used by the SMT model, and are thus being optimized, are shown in the figure, namely the following: monotone lexicalized reordering (ldm.f\_mono), language model (lm0), reverse lexical phrase translation probabilities (tm.fe\_lex), direct lexical phrase translation probabilities (tm.ef\_lex), and word penalty (wp). }
\end{figure}

We selected the most important feature weights in terms of their correlation with changes in the brevity penalty (and BLEU). The results are shown in Figure~\ref{fig:weightsel}.
As expected, the average value for the word penalty weight (dark gray solid line) is reduced
as we increase the average length of our tuning set.
This results in lower costs for longer sentences, explaining why we have higher verbosity.

However, this is not the full picture.
The monotone lexicalized reordering model (black solid line) sees significant reduction in its weight,
allowing for more reordering.
Furthermore, the weight for the direct lexical phrase translation (light gray line)
slightly increases as we increase the length of our tuning data.
This can be interpreted as increased reliance on word-to-word translations.

Thus, by changing the length of the development set,
we not only affect the word penalty, but also allow for changes in other parameters, which jointly yield better translation.\footnote{To be more precise in this analysis, more careful study needs to be done using the \emph{expected decoding cost}, i.e., multiplying the optimized weights by the mean feature values on a specific set. Nonetheless, there is no clear way to obtain such a mean feature vector without using a specific set of weights for decoding in the first place.}

\section{Conclusion and Future Work}
\label{sec:conclusion}

    We have explored the idea of customizing the tuning dataset for Statistical Machine Translation (SMT) 
    that makes the hyper-parameters of the SMT system more robust with respect to some specific deficiencies of the parameter tuning algorithms. This is an under-explored research direction, which can allow better parameter tuning. In this paper, we achieved this goal by selecting a subset of the available sentence pairs, which are more suitable for specific combinations of optimizers, objective functions, and evaluation measures.     
    In particular, we experimented with the pairwise ranking optimization (PRO) optimizer, which is known to yield too short translations.
    We have shown that the problem can be alleviated by tuning on a subset of the development dataset,
    selected based on sentence length.
    In particular, when selecting the longest 50\% of the tuning sentences,
    we achieved two-fold tuning speedup and
    competitive scores in terms of BLEU, while having a more compact dataset.
    These results rival those of alternative solutions
    that we proposed in our previous work,
    which fix the tuning-time BLEU+1 smoothing instead.
    Our analysis shows that this is due to improved parameter tuning.
    
Overall, our goal was not just to show how one can improve PRO,
but rather to draw the research attention to the more general idea of
customizing a tuning set through subset selection,
which can offer a number of 
opportunities for different experimental conditions, and more efficient training.
We believe that this is a very promising research direction, which is worth exploring further.

In the future, we plan to experiment with other language pairs and translation directions,
as well as with other optimizers such as MERT and MIRA (instead of PRO), 
and with other evaluation measures such as TER \cite{Snover06astudy}, METEOR~\cite{Lavie:2009:MMA}, and DiscoTK \cite{discoMT:WMT2014}, including also pairwise measures \cite{guzman-EtAl:2015:ACL-IJCNLP} (instead of BLEU).
We further want to study the sensitivity of these optimizer and metric combinations with respect to length
and other characteristics of the tuning dataset,
which would allow us to design targeted dataset customization strategies for them.

\section*{Acknowledgements}

We would like to thank the anonymous reviewers for their constructive comments.

\bibliography{biblio}
\bibliographystyle{acl_natbib}

\end{document}